# Impact of Cognitive Load on Human Trust in Hybrid Human-Robot Collaboration

Hao Guo [a], Bangan Wu [b], Qi Li [c], Zhen Ding [d], Feng Jiang [a,e,*], Chunzhi Yi [f,*]

[a] School of Computer Science and Technology, Harbin Institute of Technology, Harbin, China
[b] School of Management, Harbin Institute of Technology, Harbin, China
[c] School of Mechatronics Engineering, Harbin Institute of Technology, Harbin, China
[d] College of Computer and Control Engineering Northeast Forestry University, Harbin, China
[e] School of Artificial Intelligence, Nanjing University of Information Science & Technology, Nanjing, China
[f] School of Medicine and Health, Harbin Institute of Technology, Harbin, China

*Corresponding author.
E-mail addresses: fjiang@hit.edu.cn (F. Jiang), chunzhiyi@hit.edu.cn (C. Yi)

**Abstract:** Human trust plays a crucial role in the effectiveness of human-robot collaboration. Despite its significance, the development and maintenance of an optimal trust level are obstructed by the complex nature of influencing factors and their mechanisms. This study investigates the effects of cognitive load on human trust within the context of a hybrid human-robot collaboration task. An experiment is conducted where the humans and the robot, acting as team members, collaboratively construct pyramids with differentiated levels of task complexity. Our findings reveal that cognitive load exerts diverse impacts on human trust in the robot. Notably, there is an increase in human trust under conditions of high cognitive load. Furthermore, the rewards for performance are substantially higher in tasks with high cognitive load compared to those with low cognitive load, and a significant correlation exists between human trust and the failure risk of performance in tasks with low and medium cognitive load. By integrating interdependent task steps, this research emphasizes the unique dynamics of hybrid human-robot collaboration scenarios. The insights gained not only contribute to understanding how cognitive load influences trust but also assist developers in optimizing collaborative target selection and designing more effective human-robot interfaces in such environments.

# 1. Introduction

The advancement of human-robot collaboration (HRC) significantly enhances the capability of robots to support human efforts in increasing productivity. This collaborative paradigm is now widely implemented across diverse sectors such as manufacturing, healthcare, and transportation (Chen et al., 2021; Ding et al., 2016; Zhang et al., 2022), demonstrating its effectiveness in enhancing efficiency and expanding applications. For instance, a survey by the Organization for Economic Co-operation and Development (OECD) on the impact of artificial intelligence (AI) on the workplace (Lane et al., 2023) indicates that by 2022, over 30% of manufacturing processes in OECD countries will employ AI-enabled robots to boost productivity. Specifically, in industrial settings, robots are utilized to handle and install components (Buerkle et al., 2023; Segura et al., 2021), complementing human workers who perform supervisory and intervention roles in case of accidents, thereby enabling a seamless resumption of operations once issues are addressed.

Human trust in robots could play a pivotal role in optimizing HRC efficiency for several reasons. Firstly, human trust facilitates collaboration by enhancing team cohesion (Gebru et al., 2022), individuals who trust the robot's capabilities feel more confident in its contributions to collective tasks. This confidence typically results in more efficient task completion, as both humans and robots leverage their strengths in a synergistic manner. Secondly, human trust is dynamic (Braga et al., 2018; Chen et al., 2018) , varying with the task complexity and the individual's experiences with the robot.

A thorough understanding of these trust dynamics is crucial for effective task allocation and ensuring optimal performance by both parties within their capabilities. However, as the complexity of systems and scenarios escalates, the ability to establish and maintain an appropriate level of trust is constrained by the intricacies of influencing factors and mechanisms. Consequently, it is essential to investigate the factors that influence human trust in robots.

Previous research on the factors influencing human trust during human-robot interaction typically focuses on three dimensions (Khavas, 2021): task factors (e.g., cognitive load) (Desai et al., 2012), human factors (e.g., previous experience) (Ososky et al., 2013; Walters et al., 2011), and robot factors (e.g., timing of errors) (Lucas et al., 2018). Among these factors, cognitive load, which refers to the amount of mental effort required to process information, emerges as a crucial element. Firstly, cognitive load significantly influences the formation of trust (Hopko et al., 2022). Human trust hinges on the capacity to comprehend and anticipate robot actions. Under significant cognitive load, individuals might experience increased skepticism regarding their own abilities and those of the robot, thereby impacting their trust levels. Secondly, cognitive load directly affects user experience (Hopko et al., 2021). Since human trust is deeply affected by subjective feelings, excessive cognitive load can cause confusion or dissatisfaction, leading to diminished trust in robots and related technologies.

Existing literature on the relationship between cognitive load and human trust has predominantly centered on HRC involving a single operator (Boyce et al., 2015; Natarajan & Gombolay, 2020; Salem et al., 2015). In such studies, evaluations of the robot's performance often rely on its autonomous operation under human supervision or its assistance in human operations. For example, K. Lazanyi et al (Lazanyi & Maraczi, 2017)., discovered that most participants prefer to relinquish control of the autonomous car under relatively simple conditions, whereas more complex conditions tend to motivate participants to retain control over the driving task. However, these findings might not extend to hybrid HRC scenarios, such as those in stacking (Lee et al., 2021) and assembly manufacturing (Chutima, 2022). Firstly, the complexity of tasks and the coordination requirements in hybrid HRC settings are substantially higher. Unlike in single-operator environments where task steps are independent (Yang et al., 2021), hybrid HRC involves interdependent tasks, where humans and robots cooperate as equal partners, underscoring the holistic nature of the task. The interaction between cognitive load and trust in these multi-actor scenarios can differ significantly from those in single-operator contexts. Secondly, the variability of task environments poses an additional challenge. Hybrid HRC settings, particularly in manufacturing, frequently encompass a range of tasks with varying levels of difficulty and environmental conditions (Hjorth & Chrysostomou, 2022). These variations can affect cognitive load and trust differently compared to more uniform, single-operator environments. The dynamics of trust in these fluctuating settings are inherently more complex, requiring operators to continuously reassess their trust in the robot based on evolving task demands and their own cognitive capacities. Given these distinctions, our research poses the following questions: (1) How does cognitive load influence human trust in the robot within hybrid HRC tasks? (2) Does the effect of cognitive load on human trust vary across tasks of differing complexities in hybrid HRC scenarios?

In this study, we aim to explore the impact of cognitive load on human trust in robots within the context of a hybrid HRC task. Specifically, we designed an experiment where participants collaborated with a robot to construct a pyramid using blocks, across three levels of task complexity. We collected data on human trust and cognitive load, along with two performance metrics, to examine the variations following the tasks. Our findings indicate that in hybrid HRC settings, periods of high cognitive load are associated with increased trust in the robot and enhanced performance rewards. Additionally, the performance failure risk metric we developed shows a significant linear correlation with human trust in the robot during low-load and middle-load tasks.

Our research provides several key contributions. Firstly, it emphasizes the hybrid HRC scenario by introducing interdependent task steps. Specifically, we implemented a stacking task where the performance of each step influences subsequent steps, enabling humans and robots to work effectively as teammates within this hybrid HRC framework. Secondly, our findings demonstrate that varying levels of cognitive load can have diverse effects on human trust within this scenario. We manipulated cognitive load through different task complexities and measured human trust using both subjective and objective metrics to explore the relationship between cognitive load, trust, and task performance. Finally, our results facilitate the development of more effective human-robot interfaces and aid in selecting appropriate collaboration targets for hybrid HRC settings. These advancements significantly improve how humans interact with robots and enhance the overall efficiency of teamwork.

## 2. Related literature

In this section, we review the existing research on the factors influencing human trust in robots. Fig. 1 shows the position of our research in the large literature.

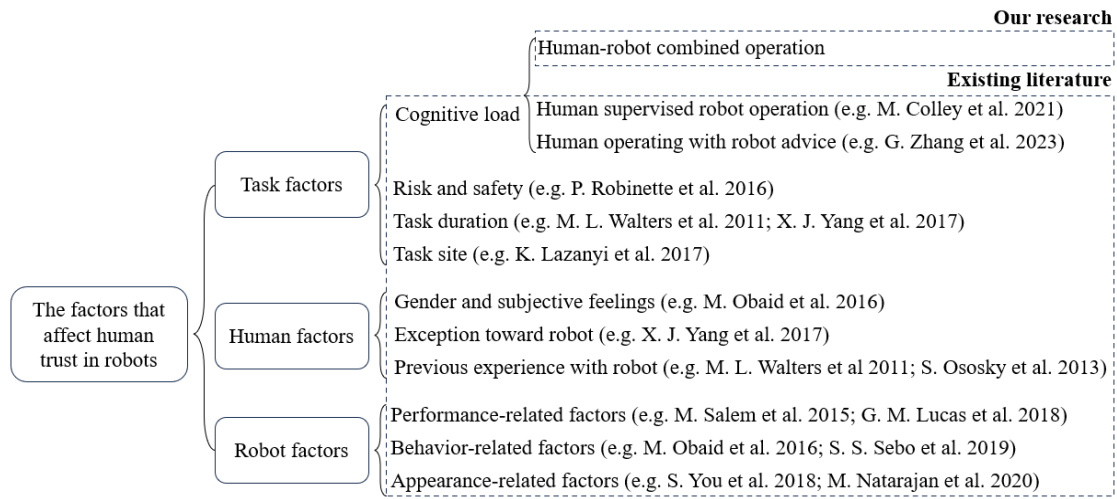

**Fig. 1** The position of our study in the literature.

### 2.1 The influencing factors of human trust in robots

Prior studies on the factors influencing human trust in robots typically categorize these factors into three groups: task factors, human factors, and robot factors (Khavas, 2021). Task factors frequently examined include human safety (Robinette et al., 2016), task site (Lazanyi & Maraczi, 2017), task duration (Walters et al., 2011; Yang et al., 2017), and cognitive load (Samson & Kostyszyn, 2015). For instance, Daronnat et al., (Daronnat et al., 2021) explored the impact of cognitive load on human trust during a robot-assisted missile command game, finding that high cognitive load detrimentally affects trust in HRC. Regarding human factors, research often focuses on variables such as gender (Obaid et al., 2016), expectations toward robots (Yang et al., 2017), and previous experience with robots (Walters et al., 2011). P. A. Hancock et al., (Hancock et al., 2011) demonstrated that these factors are crucial in forming and modulating trust during human-robot interactions. As for robot factors, studies typically assess performance-related factors (e.g., frequency of fault occurrence, timing of errors) (Desai et al., 2012; Lucas et al., 2018) and behavioral factors (e.g., naturalistic interaction, explanations after failures) (Ososky et al., 2013; Sebo et al., 2019). For example, Desai et al., (Desai et al., 2012) observed that an increase in the frequency of errors leads to a perceived decrease in robot reliability. The decline in trust can be attributed to psychological mechanisms where consistent reliability is key to maintaining trust. Therefore, frequent errors disrupt this stability, resulting in significant trust degradation.

Although numerous studies have explored the factors influencing human trust in human-robot interaction, these investigations have predominantly featured scenarios that are either human-centric HRC or robot-centric HRC (Gebru et al., 2022). In human-centric HRC, the robot's role is confined to being controlled or offering suggestions, while the human acts as the master operator (Zhang et al., 2023). Conversely, in robot-centric HRC, the robot assumes the master operator role with the human transferred to a supervisory capacity (Colley et al., 2021). Crucially, in these settings, task steps are independent, allowing the performance of the operator to be evaluated in isolation. In contrast, our study shifts the focus to a hybrid HRC, where humans and robots are co-equal operators, and the task steps are interdependent. Consequently, the performance of each operator cannot be assessed separately but must be evaluated collectively.

## 2.2 Effect of cognitive load on human trust in robots

Among the studies most relevant to our research are those examining the effect of cognitive load on human trust in robots. This architecture of work typically categorizes the interaction scenarios into two distinct types: human-supervised robot operations and robot-assisted human operations. Additionally, insights from human-human interactions indicate that increasing cognitive load generally leads to lower trust (Lopes et al., 2018; Lyons & Stokes, 2012). This finding sets a baseline for understanding trust dynamics across different interaction types, including those involving robots. In human-supervised robot operations, factors such as information overload and anxiety can negatively impact trust as cognitive load increases (Chen et al., 2016). Despite some counterexamples where increased cognitive load did not adversely affect trust (Parasuraman & Miller, 2004), this effect suggests that the human's ability to oversee and control the robot effectively diminishes under high cognitive pressures. Conversely, in robot-assisted human operations, people tend to rely more on the robot when cognitive load is high, regardless of their actual belief in the robot's suggestions (Biros et al., 2004; Gupta et al., 2020). This reliance is sometimes due to overtrust in AI-enabled functions or because the robot's design effectively matches the task (Hsu, 2002; Zhang et al., 2023). Some studies have shown that people may continue to trust robots under high cognitive load, even when they harbor doubts about the robots' advice (Biros et al., 2004; Gupta et al., 2020).

Our study distinctively diverges from existing research in terms of the scenario examined, the methods of cognitive load manipulation, and the approaches to measuring human trust. Firstly, unlike prior studies that focus on HRC scenarios where either the human or the robot primarily dominates the operation, our research employs a hybrid HRC task. We have designed a task where participants build a pyramid with blocks, requiring the human and the robot to collaboratively engage in stacking operations as equal teammates. The placement of each block crucially depends on the prior joint performance, necessitating that the combined performance of human and robot be assessed holistically. This innovative setup allows us to explore the impact of cognitive load on trust in robots more effectively. Secondly, our method of inducing cognitive load is more practical and engaging. Traditional studies often employ the N-back test (Hopko et al., 2021) to increase cognitive load, a method that can be lengthy and potentially boring, risking disengagement and negative participant responses. In contrast, we vary task complexity to naturally elevate cognitive load, ensuring participants remain appropriately challenged throughout the activity. Thirdly, we adopt both subjective and objective measures to assess human trust. While most studies rely on subjective questionnaires (Law & Scheutz, 2021), which can sometimes yield questionable reflections of trust, our study enhances measurement reliability by incorporating the dictator game (Engel, 2011; Forsythe et al., 1994). This game provides an objective measure of participants' trust levels, supplementing our subjective assessments and enriching our understanding of trust dynamics in hybrid HRC settings.

## 3. Method

To investigate the impact of cognitive load on human trust in hybrid HRC, we design an experiment where participants and a robot collaboratively build a pyramid using ten blocks. This task is structured so that the performance at each step directly influences the subsequent steps, highlighting the interdependence typical of hybrid HRC tasks. Participants undertake three sessions of the same task, each designed with varying levels of complexity to induce distinct levels of cognitive load. We assess changes in cognitive load and human trust in the robot using both pre-study and post-task questionnaires, allowing for dynamic tracking of participants' perceptions throughout the experiment. We establish two performance metrics to evaluate the effectiveness of task completion. This approach enables us to collect and analyze data on how variations in cognitive load affect participants' trust in the robot and overall performance in the hybrid HRC setting.

## 3.1 Participants

For data collection, we recruited 54 participants, ranging in age from 18 to 31, consisting of 22 females and 32 males. Prior to participation, we briefed each participant on the rules governing rewards and confirmed that payment would be made upon completion of the experiment. The compensation scheme included a fixed payment of ¥10, supplemented by an additional performance-based reward, which was contingent upon the earnings achieved during the tasks. At the conclusion of each session, we informed participants of the specific reward amount they were eligible to receive based on their performance. Following the completion of all three sessions, we calculated and disbursed the maximum reward achievable from all sessions to each participant. This reward structure was designed to ensure fair compensation for participation and to encourage optimal performance throughout the experiment.

## 3.2 Task

Participants are required to collaborate with a robot to build a pyramid using ten blocks, working through a structured task divided into ten steps, with one block being stacked per step by a robotic arm. The robotic arm is controlled either by the participants via a keyboard or by an autonomous manipulation algorithm, which performs satisfactorily in pyramid stacking tasks. To preserve a natural interaction environment, participants are not informed about the robot's evaluated performance capabilities. As depicted in Fig. 2, each task consists of three parts. In the first part, participants choose the number of blocks for the pyramid base, based on their anticipated performance. According to the pyramid stacking rule implemented in our study, as the pyramid increases in height, the number of blocks in each subsequent layer decreases by one until only one block is required. Remaining blocks are then stacked on top of the pyramid until all blocks are utilized. Participants can select a pyramid base containing between one to four blocks. The performance outcomes associated with each configuration under our stacking rules are detailed in Fig. 3.

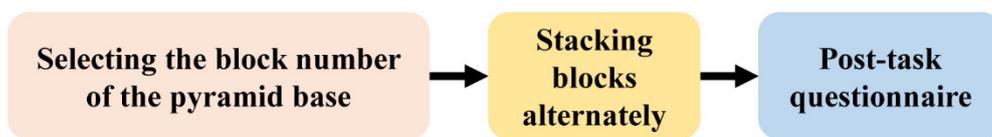

**Fig. 2** Task demonstration.

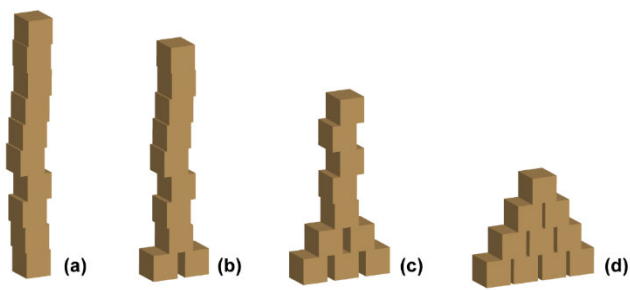

**Fig. 3** Demonstrations of the performances with different numbers of blocks in the pyramid base.

For the stacking blocks part, to clarify the task rules, we provide a concrete example as shown in Fig. 4. If the base of the pyramid is composed of four blocks and the robot is allocated seven blocks in total, red blocks represent those stacked by the robot, while blue blocks denote those stacked by the participant. The allocation of blocks, detailed in Section 3.4, sets the stage for how participants and the robot alternately manipulate the blocks. They continue this alternation until one exhausts their assigned blocks. Subsequently, the other party completes the stacking independently with the remaining blocks. Post-task questionnaires are utilized to gather data on participants' trust in the robot and their cognitive load, allowing us to assess changes in these states after each task is completed.

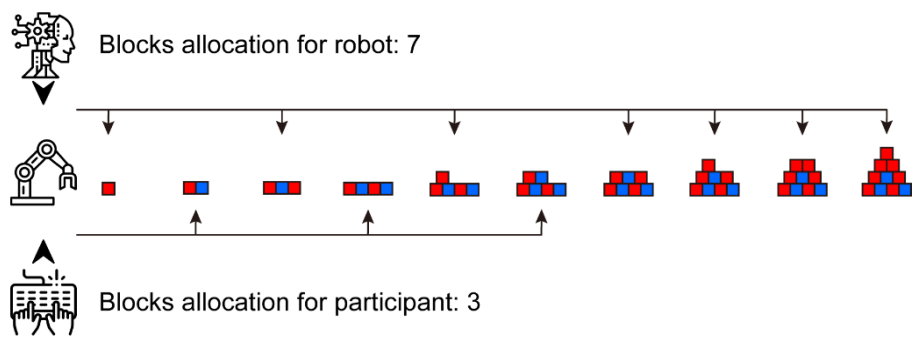

**Fig. 4** Blocks stacking demonstration.

During the task, before each step taken by the robot, participants are presented with two choices: they can either allow the robot to continue under the planned control settings or manually override the autonomous algorithm to take control. Choosing to override results in giving up any potential rewards for that step. The task proceeds until all blocks are successfully stacked into the pyramid. It is important to note that if a block is dropped before being properly placed, there is no penalty; it can simply be picked up and reattempted without affecting the overall structure of the pyramid. However, if the pyramid collapses during the stacking process, a severe penalty applies: the task is terminated, and all accumulated rewards are forfeited. This rule is designed to discourage participants from taking undue risks for greater rewards and highlights the importance of trust in the collaborative process. Additionally, if participants distrust the robot, we anticipate them taking over the task rather than risking a collapse in pursuit of higher rewards. Participants are not allowed to physically touch the blocks, ensuring that all manipulations are done through the robot to avoid any direct interference. To further minimize distractions, all blocks used are identical in appearance, negating any potential biases caused by variations in color or size. To underline the cognitive load associated with varying task complexities, each manipulation step is time-bound. Rewards for a successfully placed block are nullified if the time limit is exceeded or if participants choose to override the autonomous control, even if the placement is ultimately successful. Each point in rewards is equivalent to ¥1, which contributes to the final payment for participants.

## 3.3 Manipulation Environment

In our study, we employ the Fairino-FR16 robot (FAIR Innovation Robot Systems Co., Ltd.), which is versatile enough to be used in various industrial and healthcare scenarios[1]. As depicted in Fig. 5, this robot is equipped with a repeated positioning accuracy of ±0.03mm and six degrees of freedom. For the purposes of our task, we limit the robot's movements to three coordinate axes, which participants control via a keyboard. Visual information is critical for task execution, so two cameras are installed: one in an eye-in-hand mode to provide local visual feedback and another in an eye-to-hand mode to offer a global view of the environment. Block detection and position evaluation are conducted using the YOLO-v5 detection model (Jiang et al., 2022), which is specifically trained for our study. This model achieves an intersection ratio accuracy of 85% between the ground truth and the predicted bounding boxes, with a runtime of less than 0.3 seconds for each task step. During the task, participants are required to either directly observe the actual stacking environment or rely on image feedback from the cameras to assess the stacking situation.

---

[1] https://www.frtech.cc/APPLICATIONS

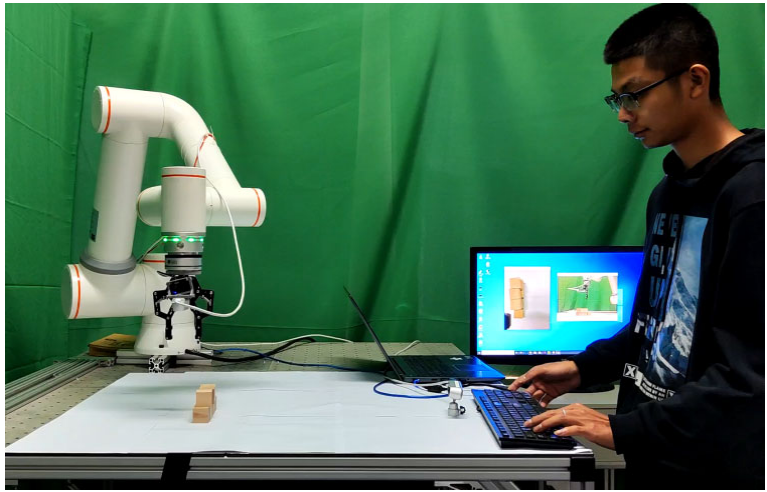

**Fig. 5** The human-robot collaboration environment.

## 3.4 Procedure

The experiment consists of three task sessions, each with varying levels of complexity: low-load, middle-load, and high-load. To mitigate the potential monotony effects associated with repeated cognitive load tasks, the sequence of these sessions is randomized for each participant. In the low-load task, as depicted in Fig. 5, participants directly observe the stacking scene without any restrictions from camera views, enabling straightforward block manipulation. For the middle-load task, shown in Fig.

6a, participants must stack blocks while relying exclusively on visual information from both eye-in-hand and eye-to-hand camera feeds displayed on the screen. This setup simulates a moderate cognitive challenge by limiting their perception to video inputs. The high-load task, illustrated in Fig. 6b, introduces an even greater cognitive load by inverting the video frames from both cameras. This inversion creates a significant disparity between the participants' visual perception and intuitive manipulation, effectively increasing their cognitive load beyond that of the middle-load tasks.

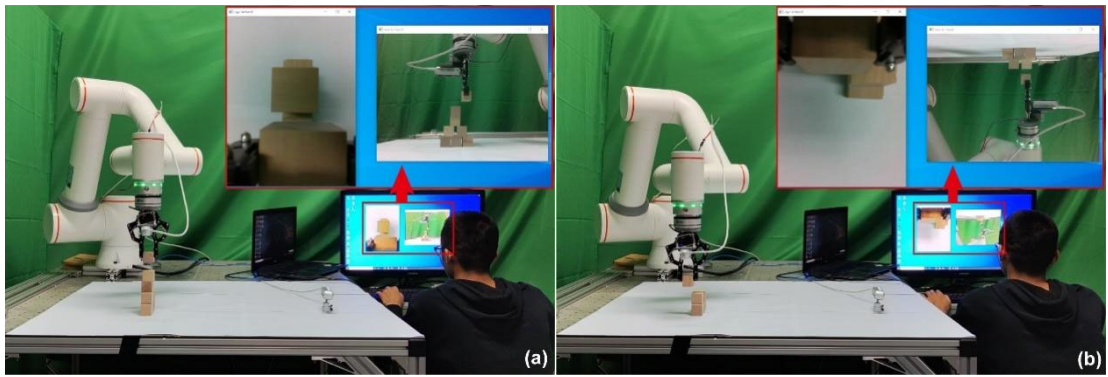

**Fig. 6** Middle-load and high-load task manipulation environments.

As illustrated in Fig. 7, the experiment procedure is divided into four parts: the preparation stage, pre-study questionnaire, task sessions, and post-study inquiry. During the preparation stage, participants view a demonstration video where a robot autonomously constructs a 2-layer pyramid using 3 blocks, with two forming the base and one placed atop them. Participants are also instructed on how to control the robot via a keyboard. Prior to beginning the tasks, participants complete the Muir and NASA-TLX pre-study questionnaires to assess their initial cognitive states. They then confirm the allocation of blocks for the robot from a total pool of 10 blocks, with the remaining blocks reserved for their own manipulation. This allocation remains consistent across all three sessions. The order of the sessions, each varying in complexity, is randomized for each participant to control for order effects. To mitigate cognitive fatigue, participants rest for five to ten minutes before each session. Upon completion of all three sessions, participants are invited to express their subjective feelings and experiences during the tasks.

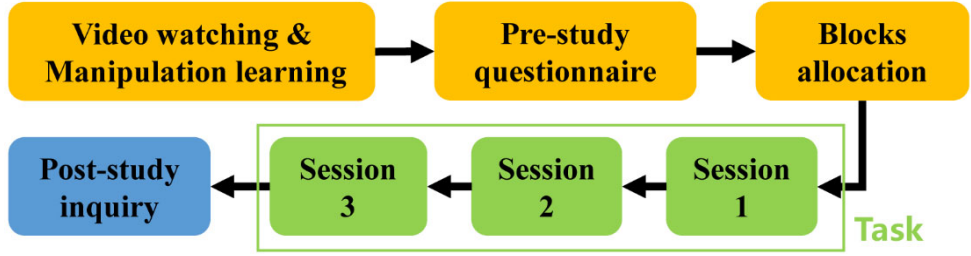

**Fig. 7** Procedure demonstration.

## 3.5 Questionnaires

In our study, participants' trust in robots and cognitive load are assessed using the Muir questionnaire (Muir, 1994; Muir & Moray, 1996) and the NASA-TLX questionnaire (Hart, 2006), respectively. Both trust in the robot and cognitive load are measured on a seven-point Likert scale, where 1 represents the lowest level and 7 the highest level of each. To align the Muir scores with the scale used in the dictator game, which ranges from 1 to 10, we adjust the collected Muir scores by multiplying them by 10/7. This conversion ensures compatibility between different measurement scales used in our study. Details of the adapted Muir questionnaire are provided in Appendix A.

## 3.6 Performance Evaluation

In this study, we develop two metrics to evaluate performance. The primary metric is the rewards associated with each task, which are calculated based on the accumulated rewards for blocks successfully stacked. Specifically, the reward for each block corresponds to the height of the layer in the pyramid where the block is placed, as detailed in Eq. (3-1). For visualization, each block in Fig. 8a is marked with a number on its corner, which signifies the reward value it represents.

$$Rewards = \begin{cases} 0, & if\ pyramid\ collapsed \\ \sum_{k=1}^{10} h(k), & other \end{cases} \quad (3\text{-}1)$$

where $k$ represents the number of blocks stacked on the pyramid, and $h$ represents the height of the layers where these blocks are placed. The height for blocks at the base layer is designated as 1, with the height incrementally increasing by 1 for each layer above the base.

In addition to rewards, we introduce failure risk as a second metric for performance evaluation. The definition of failure risk in our study is not limited to current task conditions but also incorporates historical performance data from both the robot's autonomous manipulations and the participants' manual operations. This historical context can be altered by occurrences such as bumps during stacking, which may randomly influence the stability and performance of the current block being placed. Each block is assigned a distinct Failure Risk Value (FRV), which is calculated based on its height in the pyramid and its relative position to the supporting structures beneath it. The method for calculating the FRV is detailed in Eq. (3-2).

$$FRV = abs(x_e - x_s)h \qquad (3-2)$$

where the $x_e$ represents the horizontal coordinate of the evaluated block center, $x_s$ represents the horizontal coordinate of the support center under the evaluated block, $h$ represents the height of the current block's layer. Notably, the Failure Risk Value (FRV) for blocks at the base layer is zero, reflecting their stability. If the evaluated block is within the pyramid structure where the number of blocks per layer decreases as the height increases (e.g. the part under the x-axis in Fig. 8a, blue dots denote the evaluated block centers in the pyramid part), $x_s$ is calculated as the average horizontal coordinate of the centers of the two supporting blocks below. Conversely, if the evaluated block is part of the structure above the pyramid (e.g. the part above the x-axis in Fig. 8a, red dots denote the evaluated block centers in the part stacked over the pyramid), $x_s$ corresponds to the center of the single block directly beneath it. Both $x_e$ and $x_s$ are detected and evaluated using the YOLO-v5 model, as depicted in Fig. 8b. The failure risk evaluation for the entire pyramid considers the effect of each block's placement in time sequence, according to Eq. (3-3).

$$FR = \begin{cases} FRV_{10} + \gamma FRV_9 + \gamma^2 FRV_8 + \cdots + \gamma^9 FRV_1 \\ \quad = \sum_{k=1}^{10} \gamma^{10-k} FRV_k, \quad other \\ 0, \quad if\ pyramid\ collapse \end{cases} \quad (3\text{-}3)$$

where *FR* represents the failure risk evaluation of the pyramid, *k* represents the stacking sequence of blocks, $\gamma \in (0,1)$ represents the discount factor. In this study, we focus exclusively on the final performance and failure risk results to evaluate each task, setting $\gamma$ to 0.8.

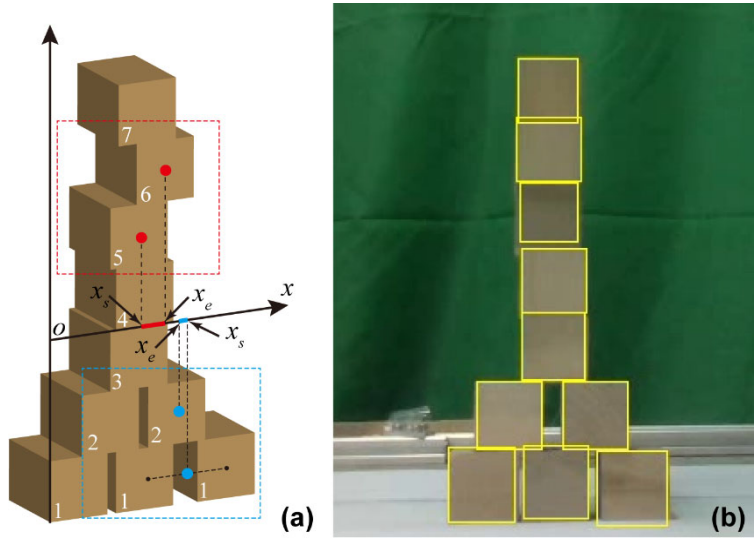

**Fig. 8** The demonstration of performance evaluation.

## 4. Results

### 4.1 The impact of cognitive load on human trust

We employ one-way ANOVA to analyze the effect of cognitive load on human trust, with the results of the trust measurements for tasks of varying complexities displayed in Figure 9. As indicated in Fig 9a, the trust level in high-load tasks is significantly higher than in low-load tasks ($p = 0.0494$). In Fig 9b, changes in self-reported trust for high-load tasks are significantly greater than those in low-load tasks ($p = 0.0262$). There are no significant differences in self-reported trust changes in middle-load tasks compared to other tasks. Averages and standard deviations of human self-reported trust in the robot for all participants are detailed in Appendix B. As shown in Fig. B.1, the perceived competence of the robot in high-load tasks is significantly higher than in the initial stage, and the dependability of the robot in high-load tasks is significantly higher than in low-load tasks. These details suggest that changes in cognitive load can induce significant increases in reliance on the robot's ability to successfully stack the pyramid.

Similarly, one-way ANOVA analyses are employed to assess behavioral trust as measured by the dictator game, revealing that human trust in the robot during high-load hybrid HRC tasks is significantly higher than both the initial stage and low-load scenarios ($p = 0.0231$ and $p = 0.0209$, respectively), as illustrated in Fig 9a. In contrast, results from middle-load tasks do not show significant differences compared to other tasks. As depicted in Fig 9b, there are notable disparities in the changes in behavioral trust, with a significant difference observed between high-load and low-load scenarios ($p = 0.005$). These results suggest that participants tend to delegate more responsibilities to the robot after experiencing high-load hybrid HRC tasks. Additionally, the results presented in Fig 9a indicate a significant discrepancy in initial trust levels between self-reported and behavioral measures ($p = 0.035$). However, no significant differences are found in the post-task results.

To further investigate the relationship between human trust and cognitive load, we conduct linear regression analyses, with the results detailed in Table 1. The analyses indicate that changes in trust are significantly correlated with changes in cognitive load under high-load conditions, as demonstrated by both the Muir questionnaire and the results from the dictator game. Conversely, this significant correlation between cognitive load and changes in trust is not observed in low-load and middle-load tasks, and mixed results in all tasks. This suggests that the highest levels of trust in hybrid HRC are associated with relatively low cognitive load states in high-load tasks.

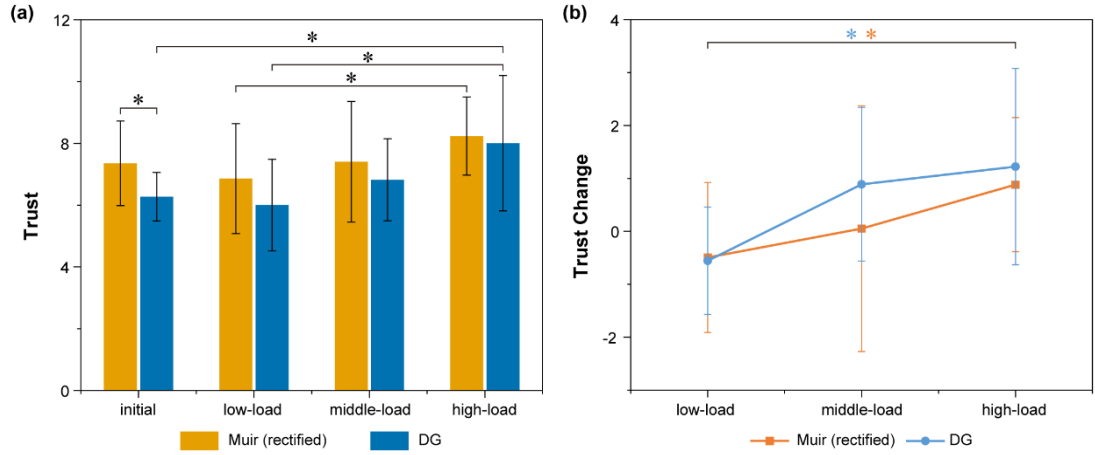

**Fig. 9** Participants' trust in the robot in different cognitive load levels.

Note. * indicates statistical significance ($p < 0.05$).

**Table 1** Linear regression results of the impact of cognitive load on human trust in the robot

| Task | All-tasks | Low-load | Middle-load | High-load |
|---|---|---|---|---|
| Muir | -0.2948 | -0.2604 | -0.7378 | -0.7114* |
|  | (0.2388) | (0.4168) | (0.5554) | (0.2406) |
| Dictator Game | 0.1116 | 0.5290 | -0.0244 | -1.3687* |
|  | (0.3017) | (0.4394) | (0.3493) | (0.5080) |
| Observation | 162 | 54 | 54 | 54 |

Note: * $p < 0.05$; ** $p < 0.01$; *** $p < 0.001$. Numbers in brackets are standard errors.

## 4.2 The impact of cognitive load on collaboration performance

As the change in cognitive load may also have an effect on collaboration performance, we further investigate the impact of changes in cognitive load on two key performance metrics: rewards and failure risk, as depicted in Fig. 10. The results show that rewards in high-load tasks are significantly higher than those in low-load tasks ($p = 0.0389$). However, rewards in middle-load tasks do not differ significantly from those in other task loads. Additionally, no significant differences are observed in failure risk across any of the tasks. Linear regression analyses exploring the relationship between cognitive load and both rewards and failure risk are presented in Table 2. These analyses reveal that changes in cognitive load do not have a significant linear correlation with either rewards or failure risk across the different tasks.

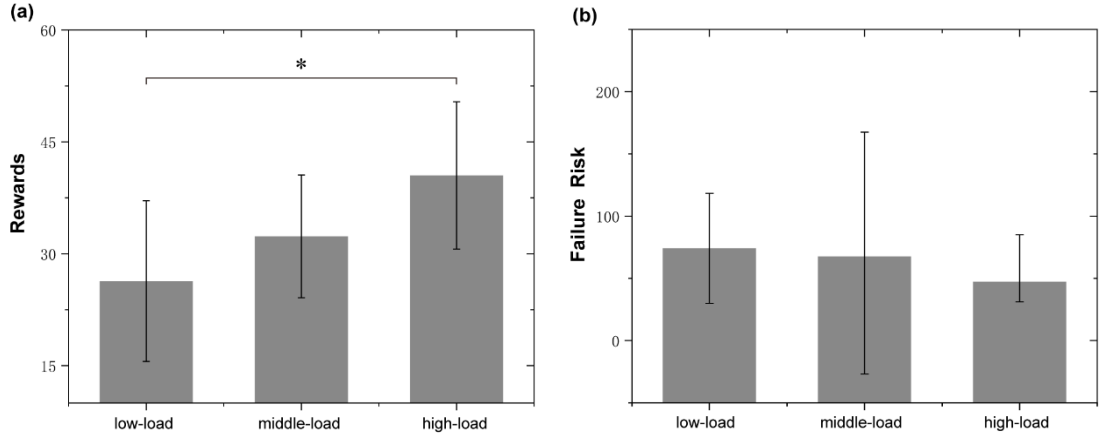

**Fig. 10** Participants' joint performance evaluation in different cognitive load levels.
Note. * indicates statistical significance ($p < 0.05$).

**Table 2** Linear regression of rewards and failure risk with cognitive load.

| Task | All-tasks | Low-load | Middle-load | High-load |
| --- | --- | --- | --- | --- |
| Rewards | 2.1775 | -2.7457 | 2.0501 | 1.9969 |
|  | (2.1310) | (4.9801) | (3.8597) | (4.6014) |
| Failure risk | -2.3928 | -7.0118 | 6.9631 | 4.1988 |
|  | (10.6740) | (21.0520) | (30.3530) | (10.3990) |
| Observation | 162 | 54 | 54 | 54 |

Note: * $p < 0.05$; ** $p < 0.01$; *** $p < 0.001$. Numbers in brackets are standard errors.

## 4.3 The relationship between human trust and collaboration performance

In previous studies, human trust and task performance have demonstrated significant correlations. In this study, we further explore the relationship between human trust and two performance metrics: rewards and failure risk. As outlined in Table 3, the first row examines the correlation between changes in human trust and rewards, while the second row assesses the impact of human trust changes on failure risk in successful tasks. The results presented in the first row indicate that rewards do not exhibit a significant linear correlation with changes in human trust across any task mode. Conversely, the findings in the second row show a significant linear correlation between failure risk and changes in human trust, particularly in low-load and middle-load tasks during successful task completion. Detailed linear regression analyses for all tasks are provided in Appendix D. Notably, a significant linear correlation between these variables is only observed in middle-load tasks.

Table 3 Linear regression of human trust change in the robot with task rewards & failure risk

| Task | All-tasks | Low-load | Middle-load | High-load |
|---|---|---|---|---|
| Rewards | 1.5135 | 5.9381 | 1.3853 | -4.6389 |
|  | (1.6941) | (3.6728) | (2.1123) | (4.3939) |
| Failure risk | -30.4930 | -35.6290* | -37.2560** | -5.2360 |
|  | (5.5676) | (9.9909) | (9.2403) | (10.488) |
| Observation | 146 | 45 | 50 | 51 |

Note: * $p < 0.05$; ** $p < 0.01$; *** $p < 0.001$. Numbers in brackets are standard errors.

## 4.4 The impact of task complexity on cognitive load

We employ one-way ANOVA to assess the impact of task complexity on cognitive load. The cognitive load at different stages, as well as changes relative to the initial stage, are depicted in Fig. 11. Specifically, Fig. 11a shows that the cognitive load in high-load tasks is significantly higher than in low-load tasks ($p = 0.041$). Furthermore, as illustrated in Fig. 11b, as task complexity increases, the change in cognitive load in high-load tasks is significantly greater than in low-load tasks. Interestingly, the average change in cognitive load for low-load tasks is negative (mean = −0.2242), suggesting a reduction in cognitive load, whereas cognitive load changes in middle-load and high-load tasks show positive values. The difference in cognitive load changes between low-load and high-load tasks is significant ($p = 0.0153$), highlighting the pronounced impact of task complexity. Detailed averages and standard deviations of cognitive load for all participants are included in Appendix B. As presented in Fig. B.1, the mental demand in high-load tasks is significantly greater than in low-load tasks ($p = 0.0258$), underscoring that increases in task complexity predominantly heighten the mental load on participants. These results underscore the direct influence of task complexities on participants' perceived cognitive load.

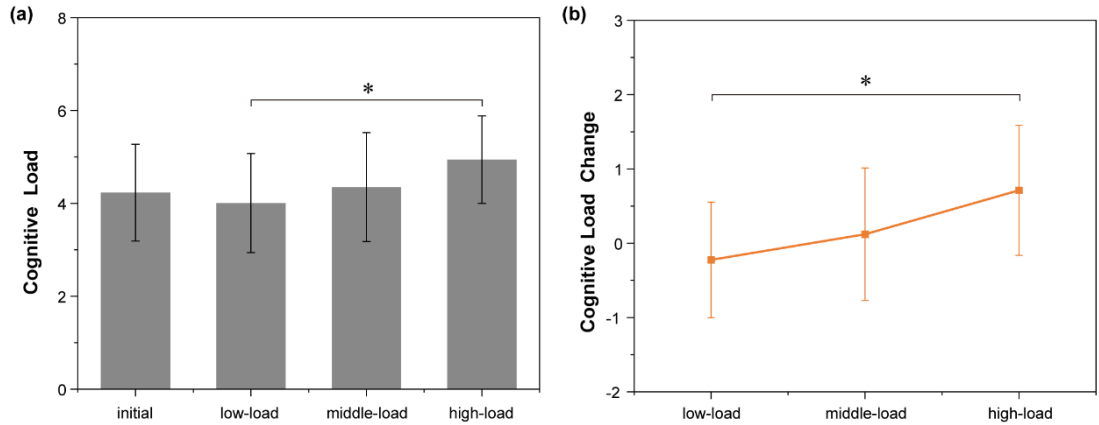

**Fig. 11** Participants' cognitive load measurement in hybrid HRC tasks of different complexities.
Note. * indicates statistical significance ($p < 0.05$).

### 4.5 Post-hoc power analysis

To evaluate the retrospective power of our study based on the sample size and the expected effect size, we conduct a post-hoc power analysis using G*Power (Faul et al., 2007). This analysis includes three tasks with a total of 54 participants. Considering the significant role of cognitive load on human trust in the robot, we assumed a medium effect size ($f = 0.25$) and set the global $\alpha$-error probability at 0.05. These input parameters result in a power of 0.8128, which is above the commonly accepted threshold of 0.8, indicating that the sample size of our study for detecting a true effect is reliable.

## 5. Discussion

### 5.1 Summary of findings

In this study, we investigate the impact of cognitive load on human trust within the context of a hybrid HRC task. The key findings are as follows: First, the cognitive load positively affects human trust, influencing both subjective self-reported trust and objective behavioral trust metrics. Second, increases in cognitive load positively correlate with enhanced performance rewards. Finally, human trust significantly reduces the performance failure risk in low-load and middle-load tasks.

Our results indicate that cognitive load significantly enhances trust in the robot. This can be attributed to two main factors: First, the task design inherently requires both participants and the robot to work integrally, making it impossible for either party to complete the task independently. Despite participants' limited familiarity with the robot's autonomous capabilities, the robot's performance remains consistent across different levels of task complexity. Under conditions of high cognitive load, participants are more likely to rely on the robot's dependable performance (Gupta et al., 2020), intervening less frequently to avoid compromising the task with their own potentially restricted capabilities. Second, our findings reveal that under high cognitive load, the time participants require to complete a task step is approximately seven times longer than that needed by the robot, often nearing the set time limits. The robot's intervention not only alleviates the time and effort burden (Dembovski et al., 2022) on participants but also boosts their expectations of the robot's effectiveness, consequently increasing their trust.

Additionally, our results indicate that cognitive load has a significantly positive impact on collaboration rewards. This finding aligns with economic decision studies cited in (Dominiak & Duersch, 2024), which demonstrate that the preference for "more certain money" is intuitive and unaffected by cognitive load. In our context, "certain money" is analogous to the rewards in low-load tasks, where a larger number of blocks at the pyramid base provides a more guaranteed reward. This assurance in low-load tasks encourages participants to pursue higher rewards in high-load tasks. Moreover, the increased trust in the robot during high cognitive load tasks leads to less frequent interventions by participants. As depicted in Fig. 11b, this results in more cautious manipulations during the stacking process, achieving a neater stack and minimizing the risk of block drops and potential collapses. These factors collectively contribute to an increase in performance rewards in high-load tasks.

Finally, in successful tasks, our results demonstrate a significant correlation between collaboration failure risk and changes in human trust in low-load and middle-load tasks. However, this correlation vanishes in high-load tasks. Two potential explanations for these findings are: First, as outlined in Section 3.6 on the failure risk value function design, the evaluation of failure risk incorporates not only current performance but also historical performance levels, applying a specific loss coefficient. In tasks where steps are interdependent, failure risk more accurately reflects the overall performance level compared to rewards alone. A high level of human trust suggests that participants are confident in the robot's capability to handle tasks under both standard and adverse conditions. This confidence tends to reduce unnecessary human interventions, allowing for more consistent robot operations throughout the task (Chen et al., 2018, 2020). Second, the absence of significant effects in high-load tasks may be attributed to participants' increased reliance on the robot, which leads them to relinquish subjective control over managing failure risk. In states of high cognitive load, as depicted in Fig. 10a, participants are inclined to chase higher rewards, prioritizing primary reward metrics over the management of failure risk (Fischer et al., 2012; Leng et al., 2021). While failure risk remains lowest in high-load tasks, this may be an incidental effect of the pursuit of high rewards rather than a direct focus on minimizing risk.

## 5.2 Implications

Our work offers significant theoretical and practical implications. Theoretically, this study enhances the existing literature on human trust in robots by emphasizing the hybrid HRC scenario with interdependent task steps (Chen et al., 2018; Soh et al., 2020; Xie et al., 2019). Unlike prior studies that primarily focus on independent and repeatable tasks where the robot acts as either the operator with human supervision or as an assistant to a human operator, our research illustrates a scenario where the human and robot function as equal partners. Each task step is interdependent, meaning the performance of one step directly influences subsequent steps. This shift highlights the importance of holistic joint performance evaluations over individual contributions, marking a significant departure from traditional HRC studies.

Besides, our work contributes to the literature on cognitive load by setting distinct task complexities to investigate heterogeneous impacts on human trust in the hybrid HRC scenario. While existing research confirms the effect of cognitive load on human trust using methods like the N-back test (Hopko et al., 2021), little is known about how these effects vary across different task complexities. We address this gap by setting distinct task complexities that maintain participants within a targeted cognitive load throughout the task. This approach allows us to investigate the relationship between cognitive load and human trust in different task complexities, providing a more detailed insights into the dynamic interplay between cognitive load and trust.

Our research also provides valuable practical implications. First, our findings inform the design of efficient human-robot interfaces. We observe that under low cognitive load, individuals prefer to take subjective initiative and execute tasks independently. This suggests that for low cognitive load scenarios, designing collaboration strategies that support and encourage human operation is beneficial. Conversely, under high cognitive load, individuals are more likely to trust and rely on the robot's autonomous operations. In such cases, implementing a strategy that leans more towards autonomy may be riskier but can significantly enhance performance and accelerate task completion, thus promoting effective ongoing HRC.

Second, our results are instrumental in optimizing collaboration target selection. The study indicates that participants exhibit higher trust in the robot and achieve neater task execution under low cognitive load. This highlights the necessity for robots in such scenarios to focus not only on the successful completion of tasks but also on the quality of performance or the product. On the other hand, in high cognitive load situations, it is crucial for robots to prioritize achieving task success, potentially at the expense of lesser focus on quality, to reduce the cognitive burden on human partners.

# 6. Conclusions

This study explores the impact of cognitive load on human trust within a hybrid human-robot collaboration scenario, characterized by interdependent task steps and collective performance outcomes. We establish that cognitive load significantly influences human trust, with its effects varying across different task complexities. Our experimental design involves collaborative pyramid building tasks where humans and robots work together under varying levels of task difficulty. To measure human trust, we utilize both subjective and objective methods: a questionnaire and a behavioral economics game. Findings reveal that human trust in the robot increases in scenarios involving higher cognitive loads compared to those with lower cognitive loads. Furthermore, performance rewards are notably higher in tasks with high cognitive load than in those with low cognitive load. Human trust significantly correlates with the performance failure risk in successful low-load and middle-load tasks. We believe this study can benefit future research on human trust and cognitive load, and help researchers develop reliable interfaces and strategies for effective human-robot collaboration.

**Appendix A. Muir and NASA-TLX questionnaires**

**Table A.1** Muir questionnaire

1. To what extent does the robot teammate perform its function properly?
2. To what extent can the robot teammate's behavior be predicted from moment to moment?
3. To what extent can you count on the robot teammate to manipulate this work?
4. To what extent does the robot teammate respond similarly to similar circumstances at different points in time?
5. Overall, how much do you trust the robot teammate?

**Table A.2** NASA-TLX questionnaire

1. How much mental and perceptual activity is required?
2. How much physical activity is required?
3. How much time pressure did you feel due to the rate or pace at which the tasks or task elements occurred?

4. How hard did you have to work (mentally and physically) with the robot to accomplish the desired level of performance?
5. How successful do you think you are in accomplishing the goals of the task set by the experimenter (or yourself)?
6. Frustration level (How insecure, discouraged, irritated, stressed, and annoyed versus secure, gratified, content, relaxed, and complacent did you feel during the task?).

## Appendix B. Means and standard deviations for human trust in robots

**Table B.1** Means and standard deviations for human self-reported trust in robot teammate

|  | Initial | | Low-load | | Middle-load | | High-load | |
| --- | --- | --- | --- | --- | --- | --- | --- | --- |
|  | Mean | STD | Mean | STD | Mean | STD | Mean | STD |
| Competence | 5.0909 | 1.1362 | 4.8182 | 1.8340 | 5.4545 | 1.4397 | 6.0909 | 0.9439 |
| Predictability | 5.5455 | 1.1282 | 5.1818 | 1.1677 | 5.6364 | 1.5015 | 6.0000 | 1.0000 |
| Dependability | 4.8182 | 1.2505 | 4.5455 | 1.4397 | 5.0000 | 1.5492 | 5.7273 | 1.1909 |
| Reliability Over Time | 5.2727 | 0.9045 | 4.9091 | 1.3003 | 4.9091 | 1.5783 | 5.5455 | 0.8202 |
| Overall Trust Degree | 5.0000 | 1.0954 | 4.5455 | 1.2933 | 4.9091 | 1.3751 | 5.4545 | 1.2933 |

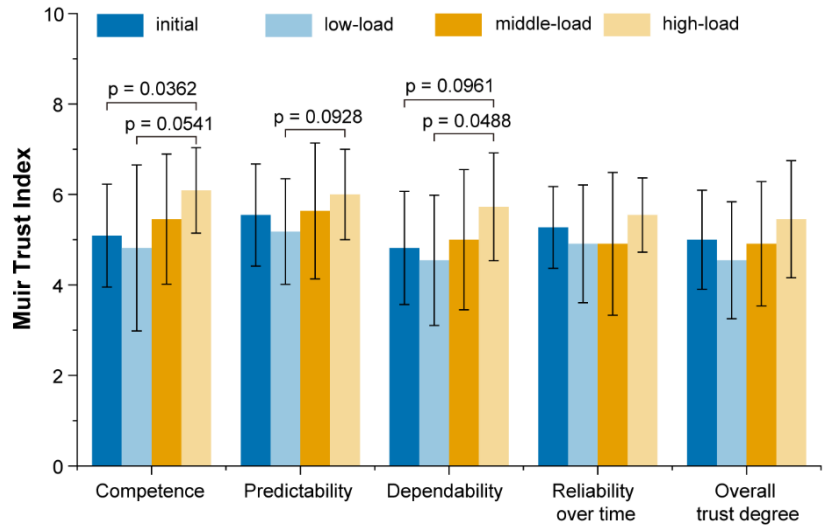

**Fig. B.1.** Means and standard deviations for trust.

## Appendix C. Linear regression analysis of the relationship between human trust in the robot and performance

**Table C.1** linear regression analysis of the relationship between human trust in the robot and performance in all tasks

| Task | All-tasks | Low-load | Middle-load | High-load |
|---|---|---|---|---|
| Rewards | 2.2355 | 6.6631 | 0.7519 | -4.9361 |
|  | (2.3193) | (5.0174) | (2.9479) | (6.0635) |
| Failure risk | -25.4460 | -15.6630 | -38.1050** | -5.6331 |
|  | (6.3717) | (15.8190) | (9.0892) | (11.4750) |
| Observation | 162 | 54 | 54 | 54 |

Note: * $p < 0.05$; ** $p < 0.01$; *** $p < 0.001$

## Appendix D. Means and standard deviations for cognitive load

**Table D.1** Means and standard deviations for cognitive load levels

|  | Initial | | Low-load | | Middle-load | | High-load | |
|---|---|---|---|---|---|---|---|---|
|  | Mean | STD | Mean | STD | Mean | STD | Mean | STD |
| Mental Demand | 3.2727 | 1.3484 | 2.8182 | 1.3280 | 3.6364 | 1.6293 | 4.1818 | 1.3280 |
| Physical Demand | 1.9091 | 0.9439 | 2.1818 | 1.4013 | 2.2727 | 1.2721 | 2.4546 | 1.2136 |
| Temporal Demand | 4.2727 | 1.4894 | 4.0909 | 1.5136 | 4.0000 | 1.6125 | 4.2727 | 1.4894 |
| Performance | 5.9091 | 1.1362 | 5.4546 | 1.4397 | 5.7273 | 1.3484 | 6.0909 | 0.8312 |
| Effort | 4.2727 | 1.1909 | 4.1818 | 1.2505 | 4.7273 | 1.3484 | 4.6364 | 1.5667 |
| frustration | 2.3636 | 1.7477 | 3.3636 | 1.1201 | 3.0909 | 1.2210 | 3.3636 | 1.3618 |
| Weighted | 4.2303 | 1.0432 | 4.2061 | 1.0677 | 4.3515 | 1.1704 | 4.6424 | 0.9392 |

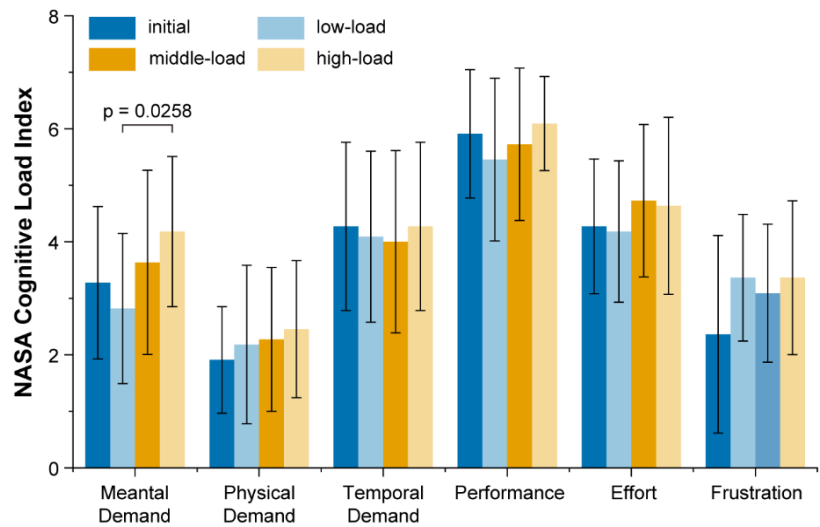

**Fig. D.1.** Means and standard deviations for cognitive load.

# Credit author statement

**Hao Guo:** Conceptualization, Methodology, Formal analysis, Software, Formal analysis, Investigation, Data Curation, Writing - Original Draft, Writing - Review & Editing, Visualization, Project administration. **Bangan Wu:** Methodology, Formal analysis, Software, Writing - Review & Editing. **Qi Li:** Software, Investigation, Data Curation. **Zhen Ding:** Investigation, Resources. **Feng Jiang:** Conceptualization, Supervision, Funding acquisition. **Chunzhi Yi:** Methodology, Writing - Review & Editing, Supervision.

# Declaration of competing interest

The authors declare that they have no known competing financial interests or personal relationships that could have appeared to influence the work reported in this paper.

# Declaration of Generative AI and AI-assisted technologies in the writing process

Statement: During the preparation of this work the authors used ChatGPT in order to improve readability and language. After using this tool, the authors reviewed and edited the content as needed and takes full responsibility for the content of the publication.

# Acknowledgements

The work of this paper is funded by the project "Human Motion Intent Sensing and Adaptation for Enhanced Human-Robot Interaction Research", which is supported by the National Natural Science Foundation of China. (No. 2018YFC0806802). Any opinions, findings, and conclusions or recommendations expressed in this material are those of the authors and do not necessarily reflect the views of the sponsor.